\pdfoutput=1

\documentclass[11pt]{article}

\usepackage{acl}

\usepackage{times}
\usepackage{latexsym}
\usepackage{booktabs}
\usepackage{nert}
\usepackage{tikz}
\usepackage{tipa}
\usepackage{enumitem}
\usepackage{adjustbox}

\usetikzlibrary{decorations.pathmorphing}

\usepackage[T1]{fontenc}

\usepackage[utf8]{inputenc}

\usepackage{microtype}

\usepackage{todonotes}

\usepackage{tcolorbox}
\newtcolorbox{mybox}[3][]{
  colframe = #2!25,
  colback  = #2!10,
  coltitle = #2!20!black,
  #1
}
\usepackage{xcolor}
\usepackage{soul}
\usetikzlibrary{positioning,fit,calc}

\title{\textsc{Jambu}: A historical linguistic database for South Asian languages}

\author{Aryaman Arora \\
  Georgetown University \\
  \eml{aa2190@georgetown.edu} \\\And
  Adam Farris \\
  Stanford University \\
  \eml{adfarris@stanford.edu} \\\AND
  Samopriya Basu \\
  Simon Fraser University \\
  \eml{samopriya\_basu@sfu.ca} \\\And
  Suresh Kolichala \\
  Microsoft \\
  \eml{suresh.kolichala@gmail.com}}

\begin{document}
\maketitle
\begin{abstract}
We introduce \textsc{Jambu}, a cognate database of South Asian languages which unifies dozens of previous sources in a structured and accessible format. The database includes 287k lemmata from 602 lects, grouped together in 23k sets of cognates. We outline the data wrangling necessary to compile the dataset and train neural models for reflex prediction on the Indo-Aryan subset of the data. We hope that \textsc{Jambu} is an invaluable resource for all historical linguists and Indologists, and look towards further improvement and expansion of the database.\footnote{The entire dataset is available at \url{https://github.com/moli-mandala/data}, and a web interface for browsing it is at \url{https://neojambu.herokuapp.com/}.}
\end{abstract}

\section{Introduction}


A particular concern of historical linguists is studying relatedness and contact between languages. Two languages are related if they share words the arose from a common source, having undergone (potentially different) regular sound changes.\footnote{Per the Neogrammarian hypothesis, sound changes are regular and \textit{exceptionless} \citep{osthoff,paul}. The reality of sound change is sometimes less ideal.} For example, the German words \textit{schlafen} and \textit{Schiff} are cognate to the English words \textit{sleep} and \textit{ship} respectively, with the German words having undergone the sound change /p/ $\to$ /f/. Using evidence like this from all of the Germanic languages, historical linguists have reconstructed the historical words that gave rise to these terms: \textit{*slāpan} and \textit{*skipą} \citep{kroonen2013etymological}.

\begin{figure}
    \centering
    \adjustbox{max width=\linewidth}{
    \begin{tikzpicture}
    [inner sep=4mm,
        place/.style={circle,draw=blue!50,fill=blue!20,thick},
        wiggle/.style={line width=1pt,draw=red,decorate,decoration={snake,post length=2mm,pre length=2mm,amplitude=0.5mm}}]
    \node at (0,3) [place,name=ancestor,label=above:$A$] {};
    \node at (-1.5,0) [place,name=childa,label=below:$B$] {};
    \node at (1.5,0) [place,name=childb,label=below:$C$] {};
    \draw [->,line width=2pt] (ancestor) -- (childa);
    \draw [->,line width=2pt] (ancestor) -- (childb);
    \draw [->,wiggle] (-2.25,0.25) -- (-0.75,3.25);
    \node at (-3.25,2.5) {\textbf{Reconstruction}};
    \node at (-3.25,2) {$p(a \mid b, c)$};
    \draw [->,wiggle] (0.75,3.25) -- (2.25,0.25);
    \node at (3.25,2.5) {\textbf{Reflex prediction}};
    \node at (3.25,2) {$p(c \mid a),~p(b \mid a)$};
    \draw [<->,wiggle] (-2,-1.5) -- (2,-1.5);
    \node at (0,-2) {\textbf{Cognate prediction}};
    \node at (0,-2.5) {$p(b \mid c),~p(c \mid b)$};
    \end{tikzpicture}}
    \caption{Three tasks of interest in computational historical linguistics. In this diagram, $A$ is the ancestor language of $B$ and $C$.}
    \label{fig:comphistling}
\end{figure}
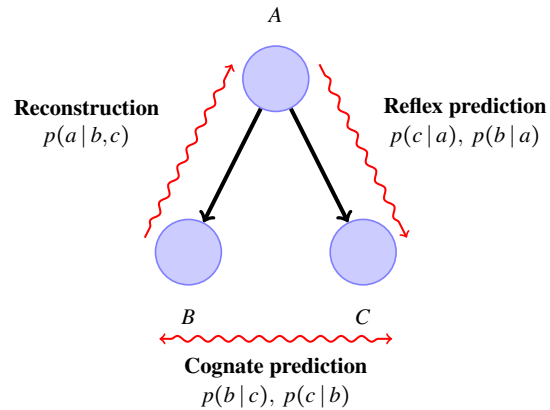

Computational historical linguistics is a growing field that seeks to apply modern computational methods to studying this kind of change \citep{comphist,list2023computational}. Massive datasets of multilingual cognates are necessary for much of the current research in this area, e.g.~on multilingual cognate detection and phoneme-level alignment \citep{list2018} and automatic comparative reconstruction of historical ancestors of languages \citep{ciobanu-dinu-2018-ab}. 

South Asia\footnote{When using the term \textit{South Asia} we refer to the Indian Subcontinent.} as a region is home to a complex historical mesh of language contact and change, especially between the Indo-Aryan and Dravidian language families \citep{masica}. Yet, South Asia is relatively understudied by linguists compared to the linguistic diversity of the region \citep{arora-etal-2022-computational}. There is no cross-family lexical dataset to facilitate computational study on South Asian historical and contact linguistics. In order to improve this unfortunate state of affairs, we introduce the \textbf{\textsc{Jambu}} cognate database for South Asian languages. \textsc{Jambu} includes all cognacy information from the major printed etymological dictionaries for the Indo-Aryan \citep{CDIAL} and Dravidian \citep{DEDR} languages, as well as data from several more recent sources. In this paper, we introduce and analyse our database and train neural models on the reflex prediction task. We hope that this resource brings us closer to the ultimate goal of understanding how the languages of South Asia have evolved and interacted over time.


\section{Related work}

\paragraph{CLDF format.} CLDF was proposed by \citet{forkel2018cross} as a standard, yet highly flexible, format for linguistic data (including cognate databases, etymological dictionaries with reconstructions, and even dictionaries). We use this format for the \textsc{Jambu} database. Many etymological databases use CLDF to effectively encode complex relations (e.g.~loaning) and metadata (e.g.~references, phonetic forms, alignments). Some which informed our database design were \citet{csd,abvd}.

\paragraph{Cognates.} \citet{batsuren-etal-2019-cognet} compiled a \textit{cognate database} covering 338 languages from Wiktionary. They noted that the meaning of \textit{cognate} varies between research communities---for our purposes as historical linguists, we prefer grouping terms with shared direct etymological sources, while much computational work (e.g.~\citealp{kondrak-etal-2003-cognates}) takes a broader definition which includes loanwords or even all semantic equivalents as cognates.

As shown in \cref{fig:comphistling}, computational historical linguistics has taken on tasks involving cognates such as automatic \textit{cognate identification} from wordlists \citep{rama-etal-2018-automatic,list2018,rama-2016-siamese}, \textit{cognate/reflex prediction}, i.e.~predicting the form of a cognate in another language based on concurrent or historical data \citep{list-etal-2022-sigtyp,khobwa,fourrier-etal-2021-cognate,marr-mortensen-2020-computerized}, and \textit{reconstruction} of the ancestor form of a cognate set \citep[\textit{inter alia}]{durham-rogers-1969-application,bouchard-etal-2007-probabilistic,ciobanu-dinu-2018-ab,meloni-etal-2021-ab,he2022neural}.

\paragraph{Other South Asian cognate databases.}  \citet{cathcart-2019-gaussian,cathcart-2019-toward,cathcart-probabilistic} and \citet{cathcart-rama-2020-disentangling} also previously made use of data from \citet{CDIAL} by scraping the version hosted online by \textit{Digitial Dictionaries of South Asia}.

There was an effort to create a new digital South Asian etymological dictionary in the early 2000s, termed the \textbf{SARVA} (South Asian Residual Vocabulary Assemblage) project \citep{sarva}. This was unsuccessful however, and only a small portion of the possible cross-family entries were complete. Our database does not incorporate it.

\section{Database}

\begin{figure*}
\centering
\small

\begin{tikzpicture}
\node(head)[draw,text width=5 cm]{454 \textbf{ápavartayati} tr. `turns away from' RV. 2. \textbf{apav\d{r}tta-} `reversed' ŚāṅkhŚr. [$\surd$v\d{r}t1]};
\node(def)[draw,text width=5 cm,below=0.15 cm of head]{1. Pk. \textit{ōvattēi} `causes to turn back'; S. \textit{oṭī} f. `turning over the edge of a cloth and hemming'; \\
2. G. \textit{oṭv$\tilde{u}$} `to hem', \textit{oṭī} f. `tucked up part of dhotī or sāṛī'...};
\node(cdial)[fit={(head) (def)}] {};
\node(arr)[right=-0.1 cm of cdial]{\huge{$\to$}};
\node(lab)[fill=orange!30,above right=1 cm and 0.1 cm of arr]{\small{OIA}};
\node(text)[draw,fill=orange!10,right=0 cm of lab]{\textit{ápavartayati}};
\node(lab05)[fill=orange!30,below=0.6 cm of lab.west,anchor=west]{\small{OIA}};
\node[draw,fill=orange!10,right=0 cm of lab05]{\textit{apav\d{r}tta-}};
\node(lab1)[fill=orange!30,below=0.6 cm of lab05.west,anchor=west]{\small{Prakrit}};
\node[draw,fill=orange!10,right=0 cm of lab1]{\textit{ōvattēi}};
\node(lab2)[fill=red!30,below=0.6 cm of lab1.west,anchor=west]{\small{Sindhi}};
\node[draw,fill=red!10,right=0 cm of lab2]{\textit{oṭī}};
\node(lab3)[fill=yellow!30,below=0.6 cm of lab2.west,anchor=west]{\small{Gujarati}};
\node[draw,fill=yellow!10,right=0 cm of lab3]{\textit{oṭv$\tilde{u}$}};
\node(lab4)[fill=yellow!30,below=0.6 cm of lab3.west,anchor=west]{\small{Gujarati}};
\node[draw,fill=yellow!10,right=0 cm of lab4]{\textit{oṭī}};
\node(cdial2)[draw,fill=black,fill opacity=0.1,fit={(lab) (lab05) (lab4) (text)}] {};
\node(arr)[left=0 cm of head]{\huge{$\leftarrow$}};
\node[above=0 cm of cdial2]{\texttt{forms.csv}};
\node(param1)[draw,left=0.1 cm of arr]{ápavartayati};
\node(paramlab1)[draw,left=0 cm of param1]{\textbf{454}};
\node(param2)[below=0.6 cm of param1.east,anchor=east]{apavahati};
\node(paramlab)[left=0 cm of param2]{\textbf{455}};
\node(param3)[above=0.6 cm of param1.east,anchor=east]{apavartana};
\node(paramlab)[left=0 cm of param3]{\textbf{453}};
\node(cdial3)[draw,fill=black,fill opacity=0.1,fit={(param1) (paramlab1) (param2) (param3)}] {};
\node(l)[above=0 cm of cdial3]{\texttt{parameters.csv}};
\node(all)[draw,draw opacity=0.3, fit={(cdial) (cdial2) (cdial3) (l)}] {};
\end{tikzpicture}
\caption{Diagram of some of the data in \textsc{Jambu} parsed from CDIAL entry 454 (\textit{ápavartayati}, `turns away from').}
\label{fig:scrape}
\end{figure*}
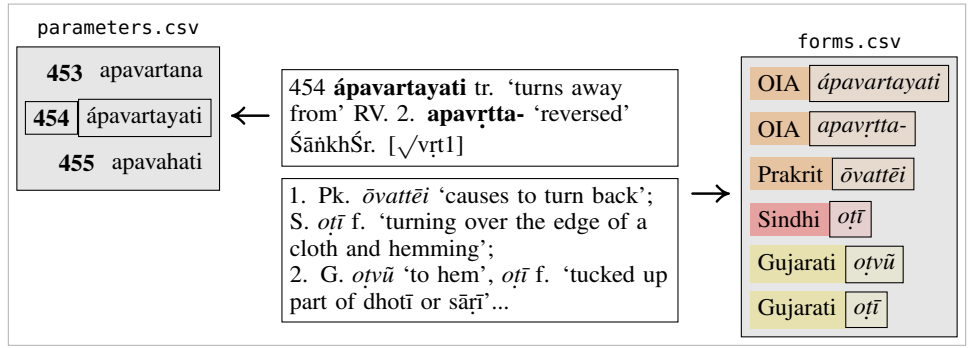

\begin{table}[t]
    \centering
    \small
    \begin{tabular}{lrrr}
    \toprule
        & \textbf{Languages} & \textbf{Cognate sets} & \textbf{Lemmata} \\
    \midrule
Indo-Aryan      &        433 &     16,464 &    194,834 \\
Dravidian       &         78 &      5,649 &     78,502 \\
Nuristani       &         22 &      3,645 &     12,088 \\
Other           &         52 &        163 &        311 \\
Munda           &         15 &        129 &      1,352 \\
Burushaski      &          2 &         41 &         48 \\
\midrule
\textbf{Total}  &        602 &     23,024 &    287,135 \\
    \bottomrule
    \end{tabular}
    \caption{Statistics about the \textsc{Jambu} database, factored by language family. \textbf{Cognate sets} counts the number of such sets that include at least one cognate from that family (and so does not sum to the total).}
    \label{tab:stats}
\end{table}

\begin{figure}
    \centering
    \includegraphics[width=\linewidth]{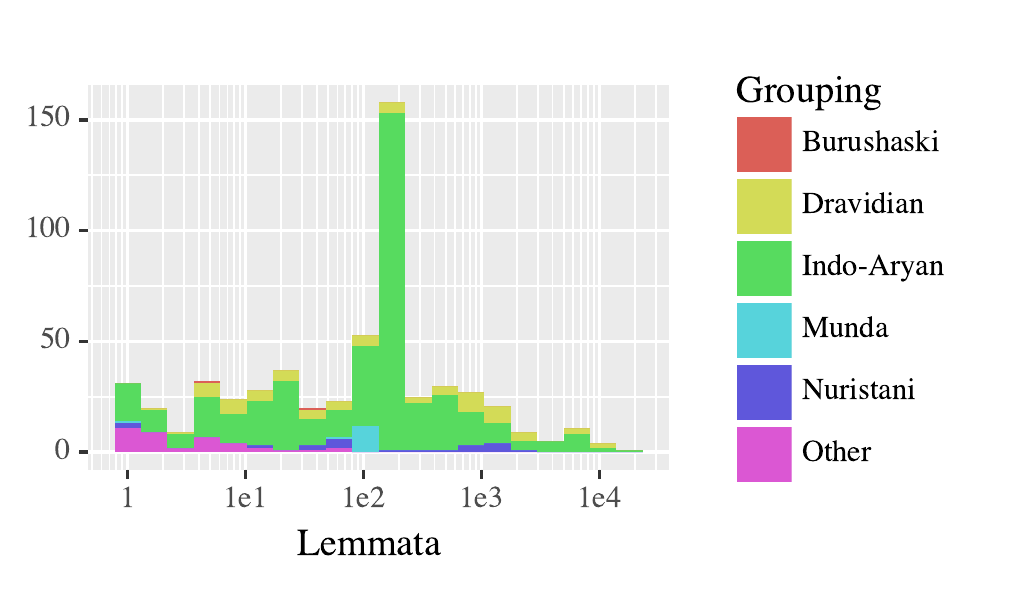}
    \caption{Distribution of languages by number of lemmata entered in \textsc{Jambu}.}
    \label{fig:lemmas}
\end{figure}

The \textsc{Jambu} database incorporates data from three major language families of South Asia: Indo-European (the Indo-Aryan and Nuristani subbranches), Dravidian, and Austroasiatic (the Munda subbranch). This comes out to 287k lemmata from 602 lects across 23k cognate sets (\cref{tab:stats}).

The data is stored in the CLDF structured data format. The overall database structure is described in the file \texttt{Wordlist-metadata.json}, which includes information about the type of data recorded in each column of each file. The file \texttt{forms.csv} includes all lemmata (word form) and associated etymological and linguistic information. The files \texttt{parameters.csv} and \texttt{cognates.csv} include all cognateset headwords and etymological notes for each. The file \texttt{languages.csv} lists all languages in the database and their geographical location. Finally, \texttt{sources.bib} lists all data references in BibTeX format.

For each lemma in \texttt{forms.csv}, we store the following information: a unique \textit{form ID}; the \textit{language ID}; the \textit{cognate set ID}, linking it to other cognate lemmata; a \textit{normalised} representation of the lemma itself, using our transcription scheme; a \textit{gloss} in English; the spelling of the lemma in the \textit{native script}; the phonemic \textit{IPA} representation of the lemma; the \textit{unnormalised} form of the lemma taken from the original source; a finer-grained \textit{cognate set ID}; \textit{notes}; and \textit{references}.

For each cognate set, we store a \textbf{headword}, which is usually a common ancestor of the words in that cognateset or a reconstruction of that ancestor if possible. We also store desiderata such as definitions and etymological notes.

Finally, we take an expansive view of what constitutes a ``language'' in our database. If a word is known to only be attested in a particular dialect, we list that dialect separately. For example, for the Shina language (northwestern Indo-Aryan), we list 32 geographical dialects. The distribution of languages by number of lemmata is depicted in \cref{fig:lemmas}.

\subsection{Data sources and scraping}
The two major data sources are CDIAL \citep{CDIAL} and DEDR \citep{DEDR}, which have been scraped in their entirety from web versions hosted by the University of Chicago's Digital Dictionaries of South Asia project.\footnote{\url{https://dsal.uchicago.edu/dictionaries/}} Since the raw data is in HTML with limited structured markup, extracting CLDF-suitable data is a significant hurdle, including matching lemmata to the appropriate language and grouping associated metadata like grammatical gender and etymological notes under the correct form (\cref{fig:scrape}). Further cleanup of data from these two sources will have to be done manually.

Since CDIAL and DEDR have not been updated in decades, we are also incorporating more recent sources that refer to them into our database, as well as etymologising newer fieldwork data manually. The additional sources we added (some partially) are listed in \cref{sec:sources}.

\subsection{Transcriptions}
One serious issue has been reconciling differing transcription systems from different sources; transcription schemes vary across sources even for the same language, since there is no standard transcription for South Asian linguistics. An illustrative example of this issue is the variable transcription of the labiodental fricative as \textit{v} or \textit{w}.

\citet{CDIAL} normalises entries from various sources into a relatively mundane Indological transcription, i.e., IAST\footnote{\url{https://en.wikipedia.org/wiki/International_Alphabet_of_Sanskrit_Transliteration}} with many extensions for the varying phonologies of South Asian languages, but not always consistently. For example, the phoneme /e\textlengthmark/ is notated $\langle\text{\={e}}\rangle$ for Sanskrit entries, but $\langle\mathrm{e}\rangle$ for Hindi (and in \citet{DEDR}, as $\langle\text{\'{e}}\rangle$ for Malto entries). Elsewhere, e.g., in Bengali and Punjabi, transcriptions adhere to the written form, which do not always adhere to any phonemic analysis of the languages in question. In the case of Kashmiri, Shina, and many other languages, there are now better analyses to base romanisation on than existed at the time of compilation of the sources of \citet{CDIAL}. Meanwhile, \citep{DEDR} does not attempt to conventionalise transcription at all, instead strictly copying the transcription from the original source; e.g.~all Bengali entries strictly reflect spelling and do not indicate the differing surface realisations of the orthographic schwa \citep{bengalischwa}.

\begin{table}[]
    \centering
    \small
    \begin{tabular}{lll}
    \toprule
    \textbf{Language} & \textbf{Original} & \textbf{Normalised} \\
    \midrule
    Old Indo-Aryan & *anug\textsubring{r}bhāyati & 
    *anug\s{r}b\super hāyati \\
    European Romani & učhar & uc\super har \\
    Shumashti & \'{\"a}šin & {\'\ae}śin \\
    Palula & beedhr\'ii & bēd\super hr\tipaupperaccent[-.1ex]{2}{\=\i} \\
    Pashai: Degano & dew'âz & dev\tipaupperaccent[-.1ex]{1}{\=a}z \\
    \bottomrule
    \end{tabular}
    \caption{Examples showing how our orthographic normalisation process affected forms from various sources.}
    \label{tab:transcrip}
\end{table}

We created a new, more rigidly standardised transcription system based on Indological conventions to unify all our data. We did not want to use pure IPA because it obscures useful cross-lingual patterns\footnote{E.g.~the Indological \textit{a} (called a schwa) varies in pronunciation across South Asia, from [\underbar{a}] (Telugu) to [\textrevepsilon] (Hindi) to [\textipa{O}\textasciitilde o] (Bengali) to [\textipa{2}] (Nepali).} and is not conventional in the Indological research community (especially considering that the database may be of use to non-linguist Indologists as well). For that reason, we use a modified IAST (for instance, using a superscript $\langle{}^\mathrm{h}\rangle$ to notate aspiration and breathy voice distinguishing these from genuine h-clusters) to suit cross-linguistic needs. Some contrasts are made more explicit while notational consistency is maintained across the board.

We used the \texttt{segments} Python library to create orthography normalisation profiles for each source's transcription scheme \citep{moran2018unicode}; some examples of the changes are shown in \cref{tab:transcrip}.
So far, forms from all source have not yet been orthographically standardised to our system. However, we developed standardisation scripts covering 204k lemmata, of which 99.7\% were automatically converted without errors.

\subsection{Web interface}
\begin{figure}
    \centering
    \includegraphics[width=\linewidth]{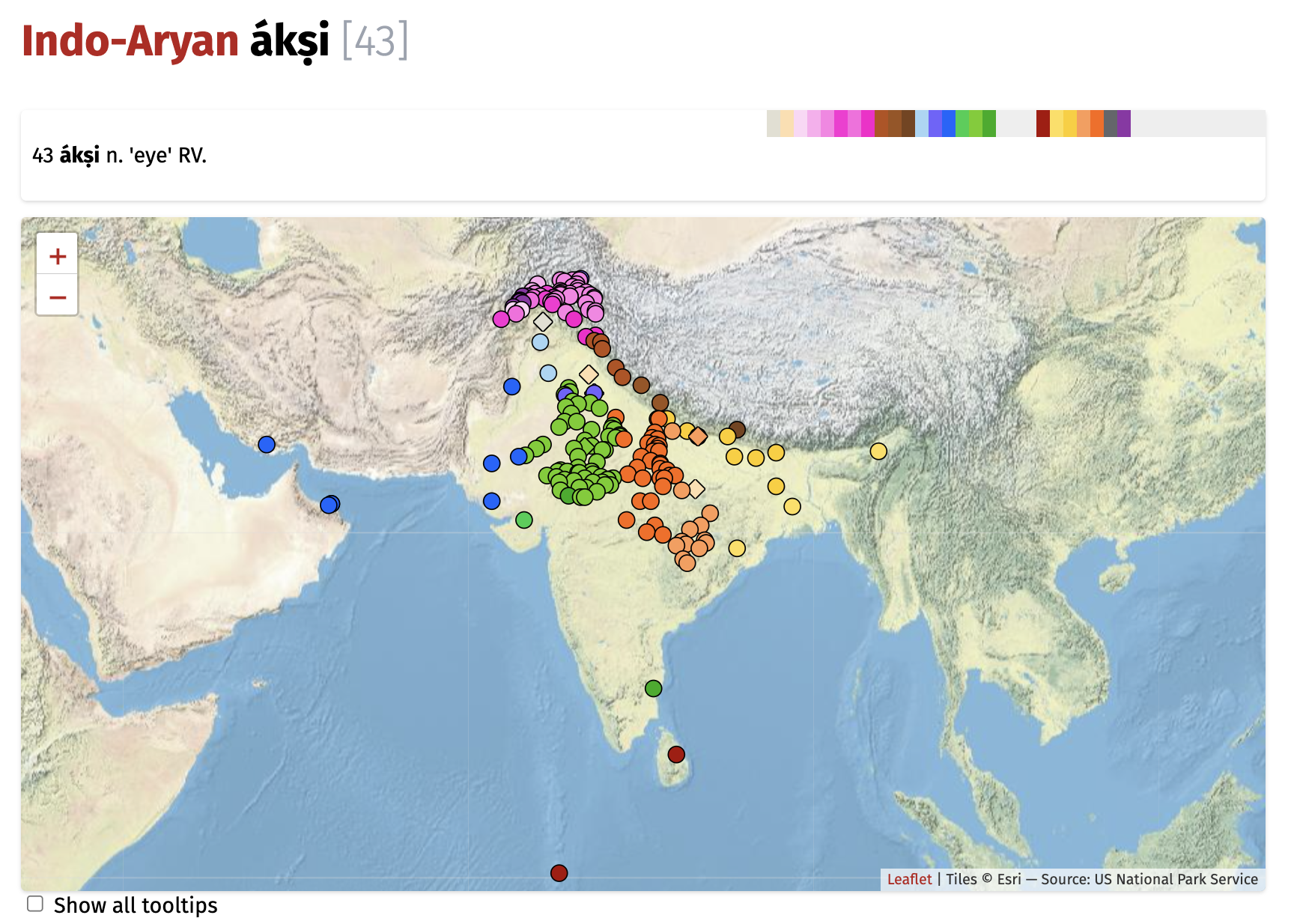}
    \caption{Web interface for Jambu, displaying reflexes of CDIAL entry 43 (\textit{ákṣi}, `eye'). See \url{https://neojambu.herokuapp.com/entries/43}.}
    \label{fig:neojambu}
\end{figure}
Finally, we developed a web interface for the \textsc{Jambu} database; see \cref{fig:neojambu}. Originally, we used the pre-existing \texttt{clld} webapp toolkit for the publication of Cross-Linguistic Linked Data\footnote{\url{https://github.com/clld/clld}}, but we later switched to a custom Flask web app designed from scratch in order to have finer control over the database structure and to execute searches on the backend more efficiently. This web interface supports search, filtering, and geographical visualisation. We hope this supersedes the unstructured search interfaces currently available for browsing older etymological dictionaries for these languages \citep{CDIAL,DEDR}.

\section{Experiment}
As a demonstration of the usability of the dataset for computational historical linguistics, we replicate the reflex prediction task of \citet{cathcart-rama-2020-disentangling}. We train neural models on the task of reflex prediction in Indo-Aryan languages, i.e.~predicting the descendant of an Old Indo-Aryan word in a given Indo-Aryan language. Rather than being restricted to data from \citet{CDIAL}, we can draw on all the sources present in \textsc{Jambu}.

We train on 80\% of the data and test on the remaining 20\%. We compare two models: a bidirectional GRU encoder-decoder with Bahdanau attention and a transformer encoder-decoder with learned positional embeddings. The optimised hyperparameters for the GRU are a learning rate of $2 \cdot 10^{-3}$, $4$ layers, and embedding and hidden size of $64$. The transformer had a learning rate of $1$ (using the parameter-based adjustment and warmup/decay schedule from \citealp{annotatedtransformer}), $3$ layers, $4$ attention heads per layer, embedding size of $64$, and FFN size of $128$. Both models were trained for $50$ epochs without early stopping with a batch size of $1024$ on a single Quadro RTX 6000, with a run completing in about 15 minutes.

We evaluate BLEU and TER on the held-out set using the SacreBLEU implementation \citep{post-2018-call}, treating a single phoneme as a `word'. Even after comprehensive hyperparameter tuning we find that both models achieve similar performances, per the results in \cref{tab:reflex}. We leave analysis of these models for future work.

\begin{table}[]
    \centering
    \small
    \begin{tabular}{lrrr}
    \toprule
    \textbf{Model} & \textbf{Perplexity} & \textbf{BLEU} & \textbf{TER} \\
    \midrule
    GRU & 2.57 & 55.91 & \textbf{34.40} \\
    Transformer & \textbf{2.53} & \textbf{56.03} & 35.15 \\
    \bottomrule
    \end{tabular}
    \caption{Performance of the two models on reflex prediction on the Indo-Aryan segment of \textsc{Jambu}.}
    \label{tab:reflex}
\end{table}

\section{Conclusion}

In this paper, we introduced \textsc{Jambu}, the largest and most up-to-date cognate database for South Asian languages. We are continuing to expand the database, incorporating all lexical data that has so far been unused in comparative linguistic work on the region. We believe that the open questions of South Asian historical linguistics cannot be resolved without examining all the information (both synchronic and diachronic) that linguists have collected about language of the region. The old etymological dictionaries are in desparate need of an update. However, much work remains. We briefly discuss some avenues of future work.

Many sources are yet to be incorporated, especially those recording loanwords from external languages (especially Persian, Arabic, English, and Portuguese) and from local literary languages (particularly Sanskrit). We have yet to distentangle cross-lectal interactions and mark lexical isoglosses, which seem necessary to understand the history of language interactions in the region; \citet{kalyan2018freeing}'s wave model of linguistic change has been thought by many scholars to be suited for South Asian languages, but it has not been operationalised yet due to a lack of comprehensive data \citep{toulmin,kogan2017genealogical}.

Another significant task ahead is extending our database structure to support indicating and analysing more complex cross-lingual interactions. For example, the database as it stands does not distinguish between inheritance from the parent language and loaning mediated by a sibling language.

We have also been working on a consistent orthography for tonemes in the languages where tones are contrastive, such as the northwestern Indo-Aryan languages \citep{baart2014tone}. Older data from these languages either does not notate tone at all (for tonality was not yet recognized, as in Gawri and Torwali), or represents it indirectly through diachronically correct, but synchronically confusing, spelling systems (as in Punjabi and Kishtwari). So, our work will also involve analyzing and incorporating new data from tonal languages, both from existing sources and our own fieldwork.

Finally, we hope to manually improve data quality once the parsing of old sources is stable. This includes fixing known mistakes, reorganising entries to better indicate indirect derivations and cross-lectal loans, and etymological notes that summarise the extant literature.

\bibliography{anthology,custom}
\bibliographystyle{acl_natbib}

\appendix
\onecolumn

\begin{figure*}[!h]
    \centering
    \includegraphics[width=\textwidth]{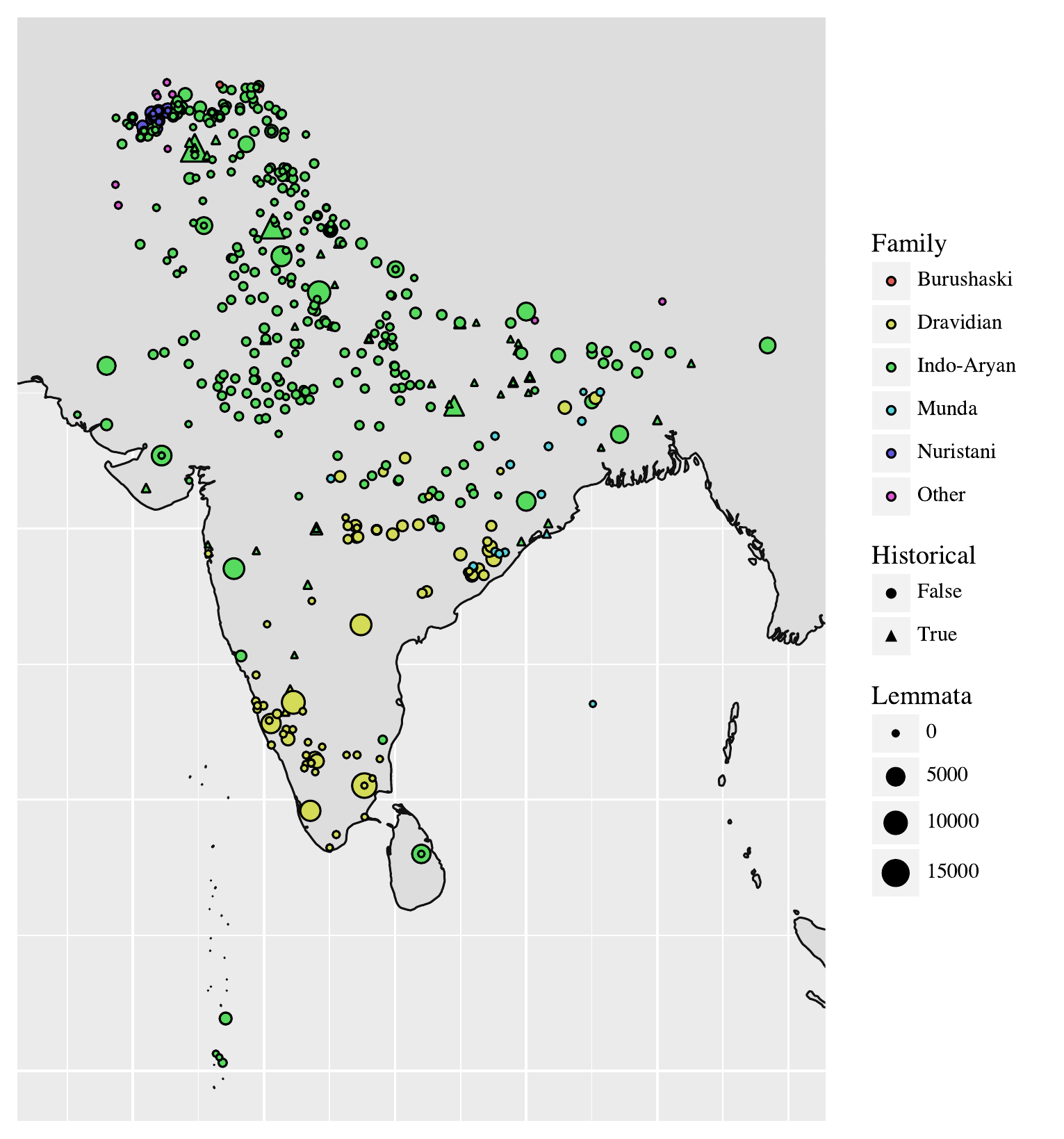}
    \caption{Map of South Asian languages present in \textsc{Jambu}, coloured by phylogenetic grouping and sized by number of lemmata included in the database. 74 lects (mostly varieties of Romani, an Indo-Aryan language, spoken in Europe and the Middle East) are not visible within the bounds of this map.}
    \label{fig:my_label}
\end{figure*}

\section{Licensing}
Data from \citet{DEDR} and \citet{CDIAL} has been scraped using the approval of the SARVA project (of which one of the authors was previously involved in) for strictly academic purposes. Additional data added to the dataset has either been manually etymologised (and therefore is an original academic contribution) or obtained with permission of the respective authors.

\newpage
\section{Other data sources}
\label{sec:sources}

\begin{table*}[!h]
\small
    \centering
    \begin{tabular}{llcc}
        \toprule
        \textbf{Language(s)} & \textbf{Reference} & \textbf{Etymologised?} & \textbf{In \textsc{Jambu}?} \\
        \midrule
        Burushaski & \citet{berger} & $\checkmark$ & $\dagger$ \\
        \midrule
        \textit{Dravidian} & \citet[DEDR;][]{DEDR} & $\checkmark$ & $\checkmark$ \\
        & \citet[DBIA;][]{emeneau1962dravidian} & $\checkmark$ & \\
        & \citet{southworth2006proto} & $\checkmark$ & $\checkmark$ \\
        & \citet{southworth2005m} & $\checkmark$ & $\checkmark$ \\
        Kurux, Malto & \citet{kobayashi2022} & $\checkmark$ & $\dagger$ \\
        & \citet{pfeiffer2018} & $\checkmark$ & $\dagger$ \\
        \midrule
        \textit{Indo-Aryan} & \citet[CDIAL;][]{CDIAL} & $\checkmark$ & $\checkmark$ \\
        Bagri & \citet{bagri} & & $\checkmark$ \\
        Bhil & \citet{bhildhule} & & \\
        Bundeli & \citet{bundeli} & & $\checkmark$ \\
        Chhattisgarhi & \citet{chattisgarhi} & & $\checkmark$ \\
        Dhivehi & \citet{fritz} & $\checkmark$ & $\checkmark$ \\
        Dogri & \citet{patyal1991etymological} & $\checkmark$ & $\checkmark$ \\
        Gawri & \citet{gawri} & & $\checkmark$ \\
        Indus Kohistani & \citet{zoller2005grammar} & $\checkmark$ & \\
        Kalkoti & \citet{liljegren2013notes} & & $\checkmark$ \\
        Kamtapuri, etc. & \citet{toulmin} & $\checkmark$ & $\checkmark$ \\
        Kannauji & \citet{kannauji} & & $\dagger$ \\
        Khetrani & \citet{elfenbein} & & $\checkmark$ \\
        Kholosi & \citet{arora2020kholosi} & $\checkmark$ & $\checkmark$ \\
        Kundal Shahi & \citet{kund} & & $\checkmark$ \\
        Kvari & \citet{jouanne} & & $\checkmark$ \\
        Maimani, Luwati & \citet{maimani} & & $\dagger$ \\
        Mandeali & \citet{patyal1982etymological,patyal1983etymological,patyal1984etymological} & $\checkmark$ & $\checkmark$ \\
        Palula & \citet{liljegren} & $\checkmark$ & $\checkmark$ \\
        Punjabi, etc. & \citet{gill} & & $\dagger$ \\
        & \citet{shackle} & $\checkmark$ & $\dagger$ \\
        Rajasthani & \citet{mewari} & & $\checkmark$ \\
        & \citet{dhundari} & & $\checkmark$ \\
        & \citet{marwari} & & $\checkmark$ \\
        & \citet{hadothi} & & $\checkmark$ \\
        & \citet{mewati} & & $\checkmark$ \\
        Shina, Domaaki & \citet{backstrom1992} & & $\checkmark$ \\
        Shina, Kashmiri & \citet{schmidt} & & $\dagger$ \\
        Thari & \citet{thari} & & $\dagger$ \\
        Tharu & \citet{boehm} & & $\checkmark$ \\
        Vaagri Boli & \citet{varma1970vaagri} & $\checkmark$ & $\dagger$ \\
        Wadiyara Koli & \citet{zubair} & & $\dagger$ \\
        Zadjali & \citet{jahdhami2017zadjali} & & $\checkmark$ \\
        \midrule
        \textit{Munda} & \citet{rau} & $\checkmark$ & $\checkmark$ \\
        & \citet{munda1968proto} & $\checkmark$ & \\
        \midrule
        \textit{Nuristani} & \citet{strand} & $\checkmark$ & $\checkmark$ \\
        \bottomrule
    \end{tabular}
    \caption{All sources included in \textsc{Jambu}, grouped together by language and family. \textbf{Etymologised?} indicates whether the original sources provided etymologies for the terms it listed; if not, we manually proposed etymologies. \textbf{In \textsc{Jambu}?} indicates what portion of the work has been incorporated into the current version of the database; $\checkmark$ means entirely while $\dagger$ means partially.}
    \label{tab:sources}
\end{table*}

\end{document}